\documentclass[10pt,twocolumn,letterpaper]{article}

\usepackage{iccv}
\usepackage{times}
\usepackage{epsfig}
\usepackage{graphicx}
\usepackage{amsmath}
\usepackage{amssymb}
\usepackage{tabularx}

\usepackage[pagebackref=true,breaklinks=true,letterpaper=true,colorlinks,bookmarks=false]{hyperref}

\iccvfinalcopy %

\def\iccvPaperID{5728} %

\begin{document}
\title{Moulding Humans:\\ Non-parametric 3D Human Shape Estimation from Single Images}

\author{
\hspace{-6mm} Valentin Gabeur$^{1,2}$ \hspace{4mm} Jean-S\'ebastien Franco$^1$   \hspace{4mm} Xavier Martin$^1$  \hspace{4mm} Cordelia Schmid$^{1,2}$   \hspace{4mm}Gr\'egory Rogez$^{3,\dagger}$ \\ 
$^1$ Inria$^*$  \hspace{20mm} $^2$ Google Research \hspace{20mm} $^3$ NAVER LABS Europe \\
}

\maketitle

\begin{abstract}
In this paper, we tackle the problem of 3D human shape estimation from single RGB images. While the recent progress in convolutional neural networks has allowed impressive results for 3D human pose estimation, estimating the full 3D shape of a person is still an open issue. Model-based approaches can output precise meshes of naked under-cloth human bodies but fail to estimate details and un-modelled elements such as hair or clothing. On the other hand, non-parametric volumetric approaches can potentially estimate complete shapes but, in practice, they are limited by the resolution of the output grid and cannot produce detailed estimates. In this work, we propose a non-parametric approach that employs a double depth map to represent the 3D shape of a person: a visible depth map and a ``hidden'' depth map are estimated and combined, to reconstruct the human 3D shape as done with a ``mould''. This representation through 2D depth maps allows a higher resolution output with a much lower dimension than voxel-based volumetric representations. Additionally, our fully derivable depth-based model allows us to efficiently incorporate a discriminator in an adversarial fashion to improve the accuracy and ``humanness'' of the 3D output. We train and quantitatively validate our approach on SURREAL and on 3D-HUMANS, a new photorealistic dataset made of semi-synthetic in-house images annotated with 3D ground truth surfaces.

\end{abstract}

\section{Introduction}

{\let\thefootnote\relax\footnote{$^*$ Univ. Grenoble Alpes, Inria, CNRS,
    Grenoble INP, LJK, 38000 Grenoble, France. \ $^\dagger$ Most of the work was
    done while the last author was a research scientist at Inria.}}

Recent works have shown the success of deep network architectures for the problem of retrieving 3D features such as kinematic joints~\cite{ChenWLSW16,Rogez2018LCR-Net++} or surface characterizations~\cite{WeiHCVL16} from single images, with extremely encouraging results.  Such successes, sometimes achieved with simple, standard network architectures~\cite{RogezS16}, have naturally motivated the applicability of these methodologies for the more challenging problem of end-to-end full 3D human shape retrieval~\cite{Bogo2016SMPLify,KanazawaBJM18}. The ability to retrieve such information from single images or videos is relevant to a broad number of applications, from self-driving cars, where spatial understanding of surrounding obstacles and pedestrians plays a key role, to animation or augmented reality applications such as virtual change rooms that can offer the E-commerce industry a virtual fitting solution for clothing or bodywear.

  \begin{figure}[tp]
	\centering
	\includegraphics[width=\columnwidth]{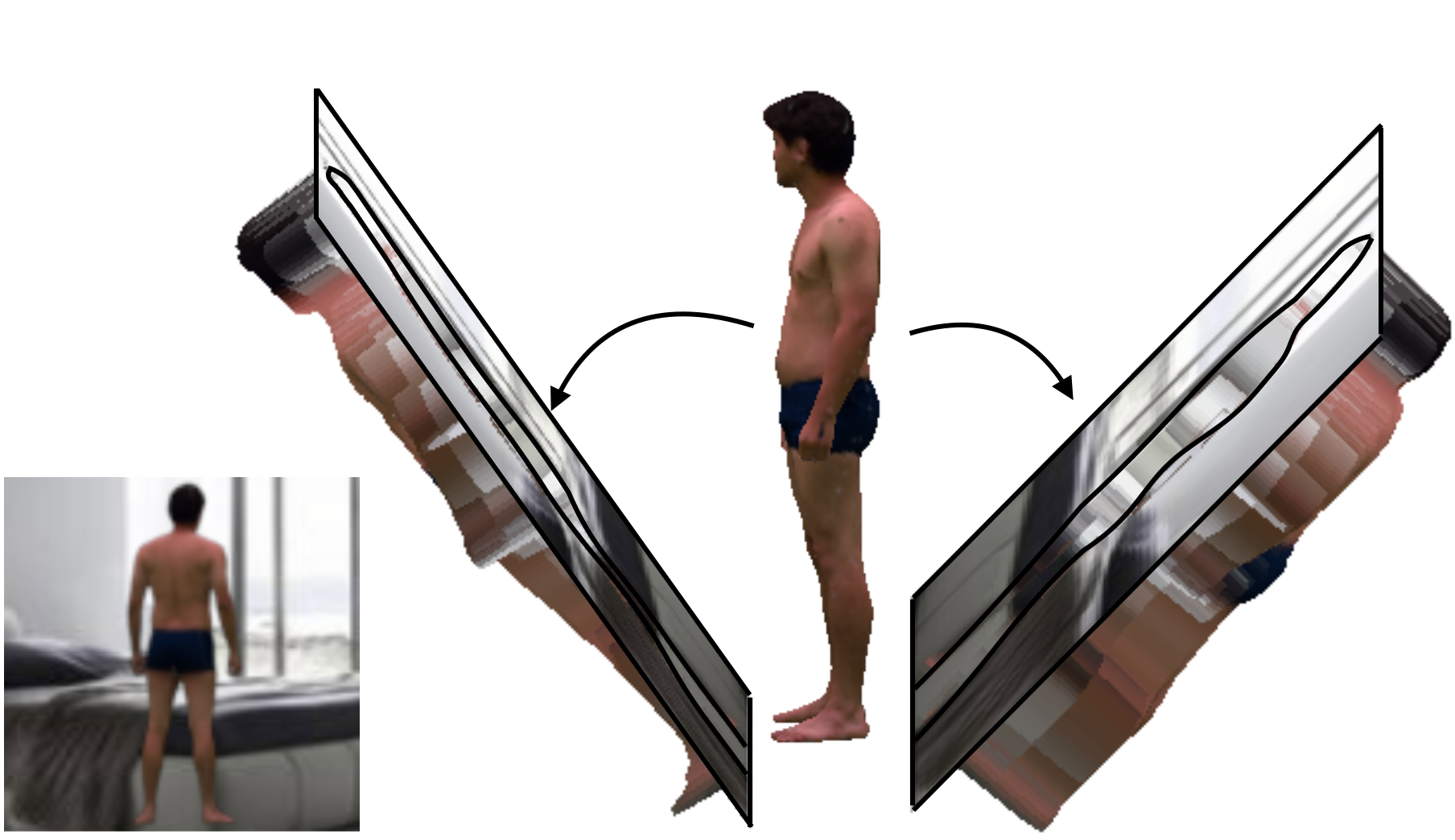} 
	\caption{Our non-parametric representation for  human 3D shape: given a single image, we estimate the ``visible'' and the ``hidden'' depth maps from the camera point of view. The two depth maps can be seen as the two halves of a virtual ``mould''. We show this representation for one of the images of our new dataset.}
	\label{splash}
	\vspace{-2mm}
\end{figure}
  \begin{figure*}[htp]
	\centering
	\includegraphics[width=\textwidth]{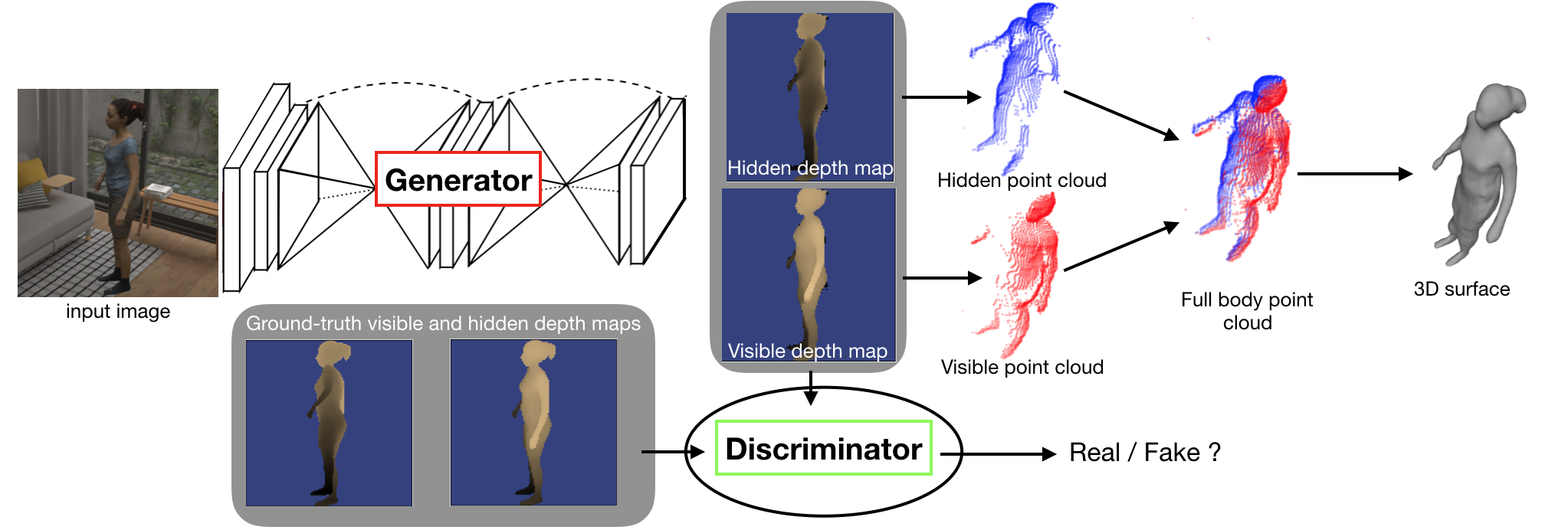} 
	\caption{Overview. Given a single image, we estimate the ``visible'' and the ``hidden'' depth maps. The 3D point clouds of these 2 depth maps are combined to form a full-body 3D point cloud, as if lining up the 2 halves of a ``mould''. The 3D shape is then reconstructed using Poisson reconstruction~\cite{Kazhdan13}. An adversarial training with a discriminator is employed to increase the  humanness of the estimation.}
	\label{pointcloud}
	\vspace{-2mm}
\end{figure*}
Designing a deep architecture that produces full 3D shapes of humans observed in an input image or a sequence of input images raises several key challenges. First, there is a representational issue. While the comfort zone of CNNs is in dealing with regular 2D input and output grids, the gap must be bridged between the 2D nature of inputs and the 3D essence of the desired outputs. One solution is to follow a parametric method and estimate the deformation parameters of a predefined human 3D model~\cite{Bogo2016SMPLify,KanazawaBJM18}. These methods are limited to the level of details covered by the model. In contrast, non parametric  approaches can potentially account for shape surface details but are prone to produce physically-impossible body shapes. This is the case of the recent volumetric approach proposed in~\cite{Varol2018BodyNet} that encodes the human body as a voxel grid whose dimensionality directly impacts the precision of the estimation. This highlights a second challenge: the dimensionality of the problem is considerably higher than what existing networks have been shown to handle, because the parametrisation sought is no longer restricted to a subset of the variability, e.g. kinematic pose of humans or body shape parameters, but to an intrinsically finer description of the body. Finally, the training data for this problem, yet to be produced, requires a particularly demanding definition and acquisition effort. The large data variability of 3D problems has motivated some initial efforts to produce fully synthetic training sets~\cite{Varol2017Surreal}, where such variability can be partially scripted. Yet recent successful methods underscore the necessity for  realistic  data, for both the general applicability of the estimation, and to keep the underlying network architecture simple, as devoid as possible of any domain specific adaptations.

 In order to  overcome  these difficulties, we propose a non-parametric approach that employs a double depth map  representation to encode the 3D shape of a person: a ``visible'' depth map capturing the observable human shape and a ``hidden'' depth map representing the occluded surface are estimated and combined to reconstruct the full human 3D shape. In this encoding of the 3D surface, the two depth maps can be seen as the two halves of a virtual ``mould'', see Figure~\ref{splash}. This representation allows a higher resolution output, potentially the same as the image input, with a much lower dimension than voxel-based volumetric representations, i.e. $O(N^{2})$ vs $O(N^{3})$. We designed an encoder-decoder architecture that takes as input a single image and simultaneously produces an estimate for both depth maps. These depth maps are then combined to obtain a point cloud of the full 3D surface which can be readily reconstructed using Poisson reconstruction~\cite{Kazhdan13}. Importantly, our fully differentiable depth-based model allows us to efficiently incorporate a discriminator in an adversarial fashion to improve the accuracy and ``humanness'' of the 3D output, especially in the case of strong occlusions.  See Figure~\ref{pointcloud}.  To train and quantitatively evaluate  our network  in near real-world conditions, we captured a large-scale dataset of textured 3D meshes that we augment with realistic backgrounds. To account for the large variability in human appearance, we took special care in capturing data with enough variability in movements, clothing and activities. Compared to parametric methods, our method can estimate detailed 3D human shapes including hair, clothing and  manipulated objects.

After reviewing the related work in Section~\ref{sec:relatedwork}, we present our two-fold contribution: our new non-parametric 3D surface estimation method is explained in Section~\ref{sec:mould} while our large-scale dataset of real humans with ground-truth 3D data is detailed in Section~\ref{sec:dataset}. Experiments are presented in Section~\ref{sec:xp} and conclusions drawn in Section~\ref{sec:conclusions}.

\section{Related Work} 
\label{sec:relatedwork}

\noindent {\bf 3D object from single images.} 
Various representations have been adopted for 3D object shape estimation.
 Voxel-based representations~\cite{Choy2016R2N2} consist in representing the 3D
 shape as an occupancy map defined on a fixed resolution voxel grid. Octree
 methods~\cite{Tatarchenko2017OGN} improve the computability of volumetric
 results by reducing the memory requirements.  Point-clouds are another widely
 employed representation for 3D shapes. In \cite{Fan2017PointSet}, Fan et al.
 estimate sets of 1024 points from single images. Jiang et al.~\cite{Jiang2018GAL} build on this idea and incorporate a geometric adversarial loss (GAL) to improve the realism of the estimations. AtlasNet \cite{Groueix2018AtlasNet} directly estimates a collection of parametric surface elements to represent a 3D shape. 
 Our representation combines two complementary depth maps aligned with the image, similar in spirit to the two halves of a ``mould'', and shares the resolution of the input image,  capturing finer details while keeping output dimensionality reasonable.
 
 Similarly to the work of Tatarchenko et al.~\cite{Tatarchenko16MultiView} on
 reconstructing vehicle images from different viewpoints, we combine the
 estimation of several depth maps to obtain a 3D shape. For human shape
 estimation, however, we work on a deformable object. Also, we focus on the visible and hidden depth maps rather than any other because of their direct correspondence with the input image.   Our two depth maps being aligned with the image, details as well as contextual image information are directly exploited by the skip connections to estimate the depth values.  Multi-views~\cite{Tatarchenko16MultiView}  do not necessarily have pixel-to-pixel correspondences with the image making depth prediction less straightforward.

\noindent {\bf 3D human body shape from images.}  
Most existing methods for body shape estimation from single images rely on a parametric model of the human body whose pose and shape parameters are optimized to match image evidence~\cite{Bogo2016SMPLify,GuanWBB09,Lassner0KBBG17,Rhodin16Contour}. This optimization process is usually initialised with an estimate of the human pose supplied by the user~\cite{GuanWBB09}  or automatically obtained through a detector~\cite{Bogo2016SMPLify,Lassner0KBBG17,Rhodin16Contour} or inertial sensors~\cite{MarcardPR16}. 
Instead of optimizing mesh and skeleton parameters, recent approaches proposed to train neural networks that directly predict 3D shape and skeleton configurations given a monocular RGB video~\cite{TungTYF17}, multiple silhouettes~\cite{DibraJOZG16}  or a single image~\cite{KanazawaBJM18, mono-3dhp2017, OmranLPGS18}. Recently,  BodyNet~\cite{Varol2018BodyNet} was proposed to infer the volumetric body shape through the generation of likelihoods on the 3D occupancy grid of a person  from a single image.

A large body of work exists to extract human representations from multiple input views or sensors, of which some recently use deep learning to extract 3D human representations~\cite{gilbert-eccv2018, Huang18ECCV, leroy2018}. While they intrinsically aren't designed to deal with monocular input as proposed, multi-view methods usually yield more complete and higher precision results as soon as several viewpoints are available, a useful feature we leverage for creating the 3D HUMANS dataset. %

More similar to ours are the methods that estimate projections of the human body: in~\cite{Varol2017Surreal}, an encoder-decoder architecture predicts a quantized depth map of the human body  while in DensePose~\cite{Guler2018DensePose} a mapping is established between the image and the 3D surface. Our method also makes predictions aligned with the input image but the combination of two complementary ``visible'' and ``hidden'' depth maps leads to the reconstruction of a full 3D volume. 
In~\cite{LunscherZ18}, the authors complete the 3D point cloud built from the front facing depth map of a person in a canonical pose by estimating a second depth map of the opposite viewpoint. We instead predict both depth maps simultaneously from a single RGB image and consider a much wider range of body poses and camera views.
All these methods rely on a parametric 3D model~\cite{Bogo2016SMPLify,KanazawaBJM18,Lassner0KBBG17} or on training data annotated~\cite{Guler2018DensePose} or synthesised~\cite{Varol2017Surreal} using such a model. These models of humans built from thousands of scans of naked people such as the SMPL model~\cite{LoperM0PB15} lack realism in terms of appearance. We instead propose to tackle real-world situations, modeling and estimating the detailed 
3D  body shape including clothes, hair and  manipulated objects.%

\noindent {\bf 3D human datasets.} 
Current approaches for human 3D pose estimation are built on  deep architectures trained and evaluated on large datasets acquired in controlled environments with Motion Capture systems~\cite{CMUposedataset,IonescuPOS14,SigalBB10}. However, while the typology of human poses on these datasets captures the space of human motions very well, the visual appearance of the corresponding images is not representative of the scenarios one may find in unconstrained real-world images. There has been a recent effort to generate in-the-wild data with ground truth pose annotation~\cite{MehtaSMXSPT18,Rogez2017LCR-Net}. All these datasets provide accurate 3D annotation for a small set of body keypoints and ignore 3D surface with the exception of~\cite{Lassner0KBBG17} and~\cite{vonMarcard2018} who annotate the SMPL parameters in real-world images manually or using IMU. Although the resulting dataset can be employed to evaluate  under-cloth 3D body shapes, its annotations are not detailed enough, and importantly, its size is not sufficient to train deep networks.

To compensate for the lack of large scale training data required to train CNNs, recent work has proposed to generate synthetic images of humans with associated ground truth 3D data~\cite{ChenWLSW16,RogezS18,Varol2017Surreal}. In particular, the Surreal dataset~\cite{Varol2017Surreal}, produced by animating and rendering the SMPL model~\cite{LoperM0PB15} on real background images, has proven to be useful to train CNN architectures for body parts parsing and 2.5D depth prediction~\cite{Varol2017Surreal}, 3D pose estimation~\cite{RogezS18,Rogez2018LCR-Net++}, or 3D shape inference~\cite{Varol2018BodyNet}. However, because it is based on the SMPL model, this dataset is not realistic in terms of clothing, hair or interactions with objects and cannot be used to train architectures that target the estimation of a detailed 3D human shape. We propose to bridge this gap by leveraging multi-camera shape data capture techniques~\cite{Casas14,Vlasic2008}, introducing the first large scale dataset of images showing humans in realistic scenes, i.e. wearing real clothes and manipulating real objects, dedicated to training with full 3D mesh and pose ground-truth data. Most similar to ours are the CMU Panoptic dataset~\cite{Joo_2015_ICCV} that focus on social interactions and the data of ~\cite{YangFHW16} that contains dense unstructured geometric motion data for several dressed subjects.

\section{Methodology}
\label{sec:mould}

In this section, we present our new non-parametric 3D human shape representation and detail the architecture that we designed to estimate such 3D shape from a single image.

\subsection{``Mould'' representation}
 
We propose to encode the 3D shape of a person through a double 2.5D depth map representation: a ``visible'' depth map that depicts the elements of the surface that are directly observable in the image, and a ``hidden'' depth map that characterises the occluded 3D surface. These two depth maps can be estimated and combined to reconstruct the complete human 3D shape as done when lining up  the two halves of a ``mould''. See example in Figure~\ref{pointcloud}. 

Given a 3D mesh, obtained by animating  a 3D human model or by reconstructing a real person from multiple views, and given a camera hypothesis, i.e. location and parameters,  we define our two 2D depth maps $z_{vis}$ and $z_{hid}$ by ray-tracing. Specifically, we cast a ray from the camera origin, in the direction of each image pixel location $(u,v)$ and find the closest intersecting point on the mesh surface:
\begin{align}
 z_{vis}[u,v] = \min_{k \in \text{Ray}(u,v)}||{\bf p}_k||_2 \label{eq:depth} 
\end{align}
 for the visible  map, and the furthest one for the hidden  map:
\begin{align}
 z_{hid}[u,v] = \max_{k \in \text{Ray}(u,v)}||{\bf p}_k||_2, \label{eq:depth2} 
\end{align}
\noindent where 3D points $\{{\bf p}_{i}\}=\{(p_{x,i},p_{y,i},p_{z,i})\}$ are expressed in camera coordinate system and the L2-norm $||.||_2$ is  the distance to the camera center. $\text{Ray}(u,v)$ denotes the set of points ${\bf p}_{i}$ on the ray passing through pixel $(u,v)$ obtained by hidden surface removal and visible surface determination.

To be independent from the distance of the person to the camera,  we center the depth values on the center of mass of the mesh, i.e. $z_{orig}: z_{vis}[u,v] '=z_{vis}[u,v] -z_{orig} \; \forall u,v$, and similarly for $z_{hid}[u,v]$.
Since they are defined with respect to the same origin, the 2 depth maps $z_{vis}[u,v]$ and $z_{hid}[u,v]$ can be readily combined in 3D space by merging their respective 3D point clouds into a global one:
\begin{align}
\{{\bf p}_{i}\}=\{{\bf p}_{i}\}_{vis} \bigcup \; \{{\bf p}_{i}\}_{hid}
\end{align}
An example of such a point cloud is depicted in Figure~\ref{pointcloud}, where points corresponding to  $z_{vis}[u,v]$ and $z_{hid}[u,v]$ are respectively colored in red and blue.
In practice,  to keep the depth values within a reasonable range and  estimate
them more accurately, we place a flat background a distance $L$ behind the
subject to define all pixels values in the depths maps in the range $[-z_{orig}
\dots L]$. Points ${\bf p}_{i}$ of the point clouds are then selected as belonging to the human surface  if $p_{z,i}\leq L-\epsilon$.

As in volumetric representation through voxel grid, our method also encodes 3D surfaces and point clouds of diverse sizes into a fixed size representation, making a 3D surface easier to consider as a deep network target. However, in our case, we can work at the image resolution with a much lower output dimensionality $O(N^{2})$ than voxel-based volumetric representations $O(N^{3})$,  N being the size of the bounding box framing the human in the input image.
 
\begin{figure}[tp]
	\centering
	\includegraphics[width=\columnwidth]{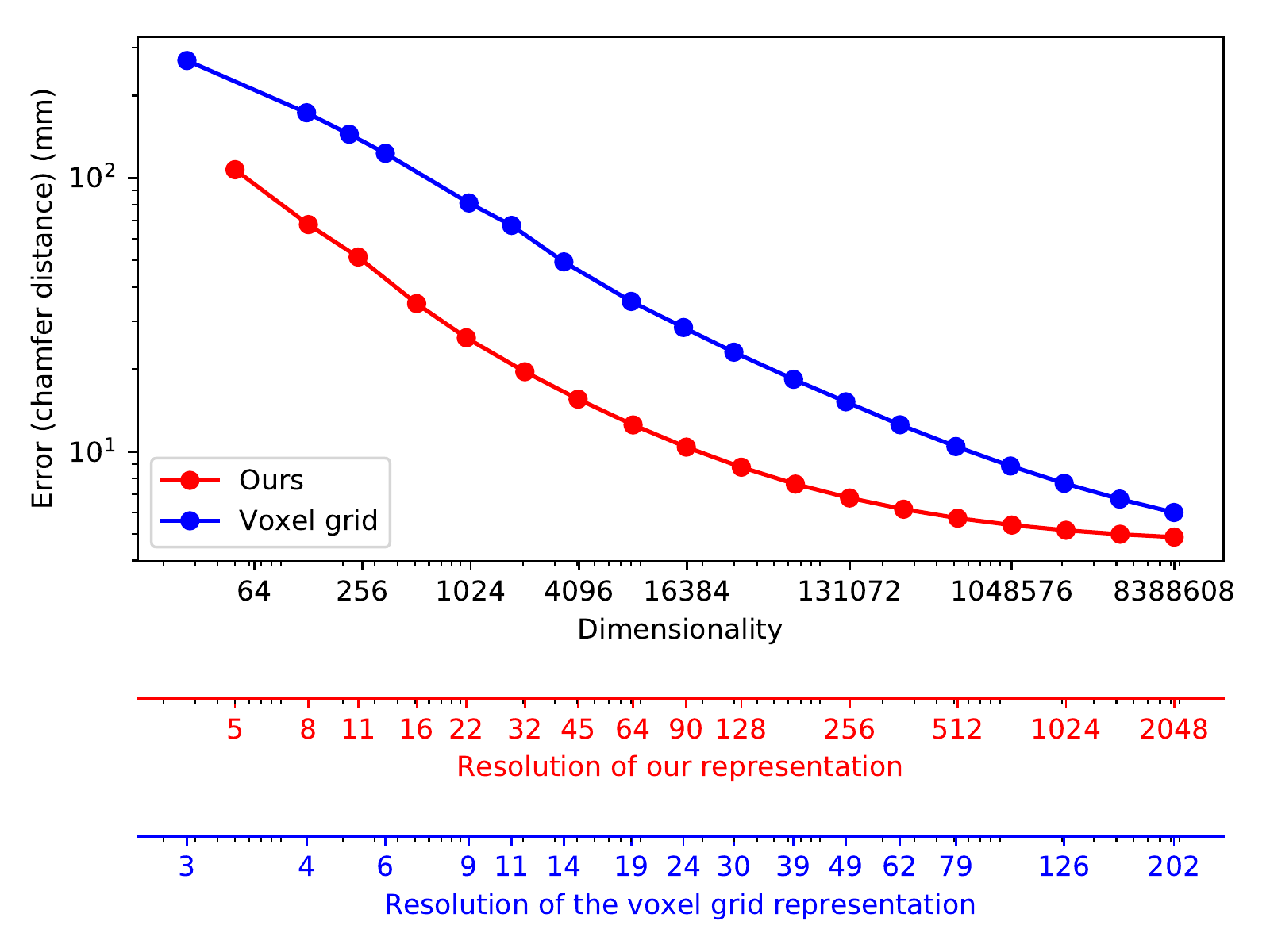} 
	\caption{Reconstruction error for voxel grid and our ``mould'' when augmenting the dimensionality D of the representation, D=$N^3$ for voxels grid and D=$2N^2$ for ours.}
	\label{representation}
\end{figure}

 We numerically validated the benefit of our representation compared to a  voxel grid approach by encoding a random set of 100 meshes (picked from our 3D HUMANS dataset presented in Section~\ref{sec:dataset}) at different resolutions and computing the 3D reconstruction error (average Chamfer distance) between ground-truth vertices and the resulting point clouds. This comparison is shown in Figure~\ref{representation}.  The error obtained with our mould-representation decreases and converges to a minimum  value that corresponds to surface details that cannot be correctly encoded even with high resolution depth maps, i.e. when some rays intersect more than twice with the human surface for particular poses. In practice, we show in Section~\ref{sec:xp} that this can be solved by employing a Poisson reconstruction to obtain a smooth 3D surface, including  those areas.  We can extrapolate from Figure~\ref{representation} that voxel grids can reach perfect results with an infinity of voxels, but for manageable sizes, our representation allows to capture more details.

\subsection{Architecture}

We formulate the 3D shape estimation problem as a pixel-wise depth prediction task for both visible and hidden surfaces.  Our framework builds on the stacked hourglass network proposed by Newell et al.~\cite{NewellY16StackedHourglass} 
that consists of a sequence of modules shaped like an hourglass, each taking as input the prediction from the previous module. Each of these modules has a set of convolutional and pooling layers that process features down to a low resolution and then upsample them until reaching the final output resolution. This process, combined with intermediate supervision through skip connections, implicitly captures the entire context of the image. Originally introduced for the task of 2D pose estimation and employed later for  part segmentation and depth prediction~\cite{Varol2017Surreal}, this network is an appropriate choice as it predicts a dense pixel-wise output while capturing spatial relationships associated with the entire human body.  
 
We designed a 2-stack hourglass architecture that takes as input an RGB image $I$ cropped around the human and outputs the 2 depths maps $z_{vis}$ and $z_{hid}$ aligned with $I$. We use a $\mathcal{L}_{L1}$ loss function defined on all pixels of both depth maps. 
The loss function to be minimized is thus the average distance between the ground truth $z_{p}$ and the estimation $\widehat{z_{p}}$:
\begin{equation} \label{L1 Loss}
\mathcal{L}_{L1}=\frac{1}{P}\sum_{p=1}^{P}|z_{p}-\widehat{z_{p}}|,
\end{equation}
with $P$ being the number of pixels in the batch and $\widehat{z_{p}}$ the network output for pixel $p$, including pixels in both $z_{vis}[u,v]$  and $z_{hid}[u,v]$ maps.
 
We also experimented with an $\mathcal{L}_{L2}$ loss but found that it overly penalizes outliers, i.e. pixels incorrectly assigned to background and vice versa, and therefore focuses only on that task. By using the $\mathcal{L}_{L1}$ norm, we force the network to not only segment the image correctly, i.e discriminate the subject from the background, but also provide an accurate estimation of the depth at each pixel. 

   \begin{figure*}[htp]
	\centering
	\begin{tabular}{c|c}
	\includegraphics[clip, trim=1cm 7cm 0cm 4cm, height=0.3\textwidth]{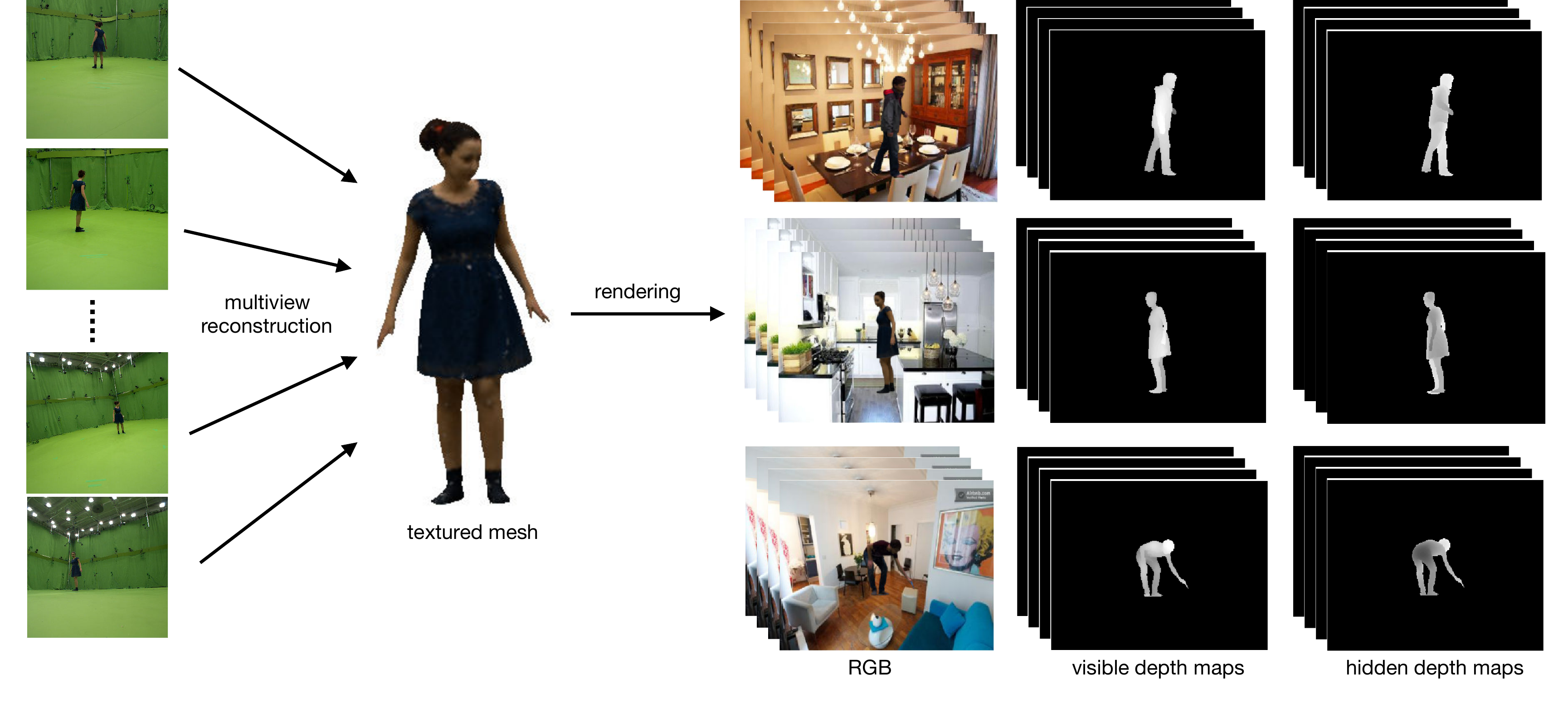} &
	 \includegraphics[height=0.35\textwidth]{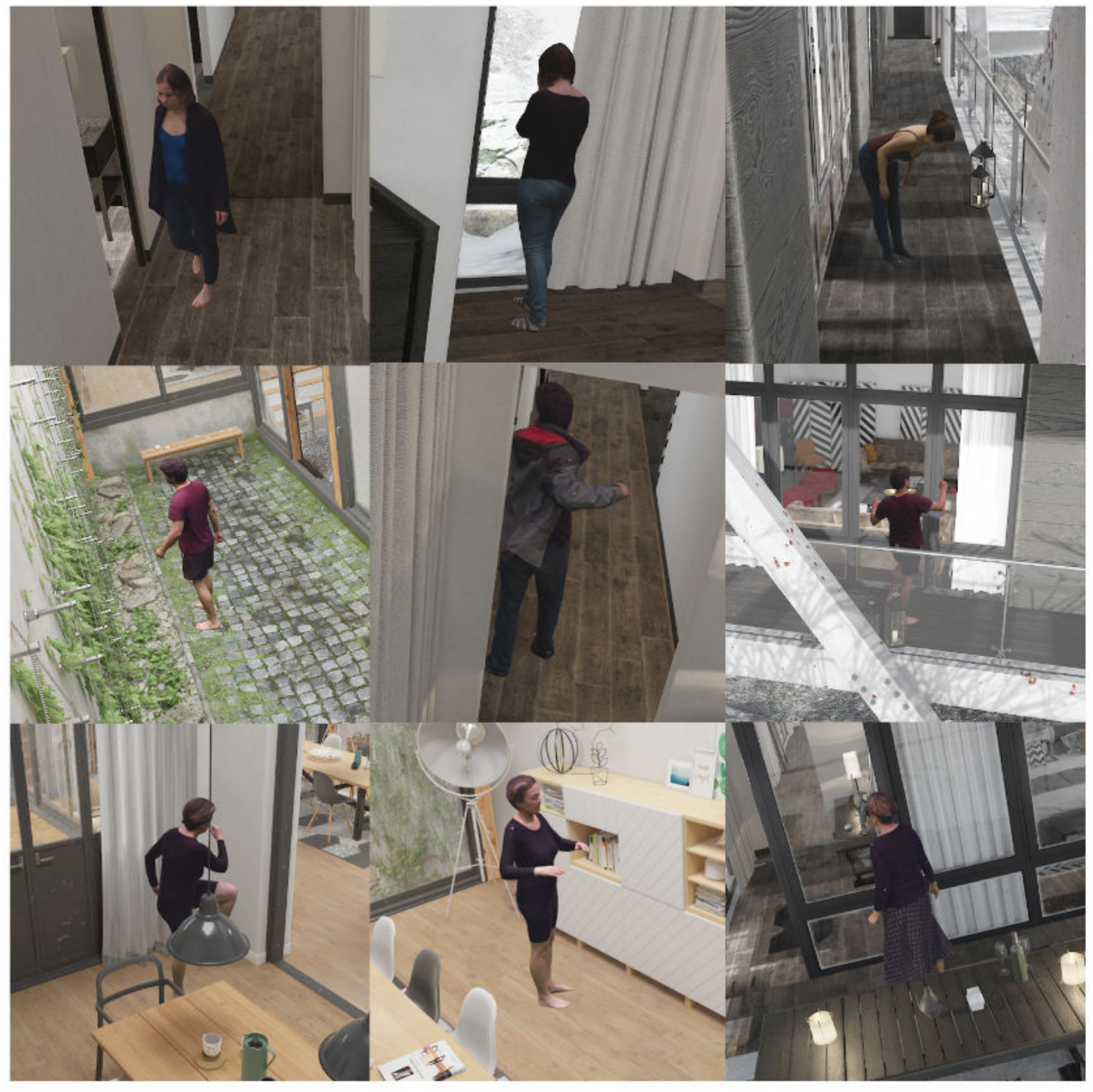} \\
	 (a)&(b)
	\end{tabular}
	\caption{Data generation. (a) We captured 3D meshes of humans, wearing real clothes, moving and manipulating objects using a multi-camera platform. We  then rendered these models on real-world background scenes and computed ground-truth visible and hidden depths maps. (b) We also generated a  test set by rendering our meshes on realistic 3D environments.}
	\label{dataset}
	\vspace{-5mm}
\end{figure*}

\subsection{Adversarial training}

As observed with other non-parametric methods~\cite{Varol2018BodyNet} but also with approaches relying on a model~\cite{KanazawaBJM18}, our network  can sometimes produce implausible shapes that do not look human, especially when a limb is entirely occluded by other parts of the body. To improve the accuracy and the ``humanness'' of our prediction, we follow an adversarial training procedure inspired by the Generative Adversarial Networks (GAN) ~\cite{Goodfellow14GAN}.  Our fully derivable depth-based model allows us to efficiently incorporate a discriminator in an adversarial fashion, i.e., the goal for the discriminator will be to correctly identify ground truth depth maps from generated ones.  On the other hand, the generator objective will be two-fold: fitting the training set distribution through the minimization of the $\mathcal{L}_{L1}$ loss (Equation~\ref{L1 Loss}) and tricking the discriminator into classifying the generated depth maps as ground truth depth maps through the minimization of the $\mathcal{L}_{GAN}$ loss:
\begin{equation}
\mathcal{L}_{GAN}(G,D) = E_{I,z}[\log D(I,z)] + E_I[\log (1 - D(I,G(I))].
\end{equation}\
Our discriminator $D$ will be trained to maximize the $\mathcal{L}_{GAN}$ loss by estimating 1 when provided with ground-truth depth maps $z$ and estimating 0 when provided with generated depth maps $G(I)$. %
In order to weigh the contribution of each loss, we will use a factor $\lambda$, our full objective being modeled as a minimax game:
\begin{equation}
(G^*, D^*) = \arg \min_G \max_D (\mathcal{L}_{GAN}(G,D) + \lambda \mathcal{L}_{L1}(G)).
\end{equation}\
The $\mathcal{L}_{L1}$ loss will be used to learn the training set distribution by retrieving the low-frequency coefficients while the $\mathcal{L}_{GAN}$ loss will entice the generator into predicting realistic and precise depth maps. It is important to note that the discriminator is only used to guide the generator during the learning. The discriminator is not used at test time.

The architecture employed as our discriminator is a 4 stack CNN. Each stack is composed of a convolutional layer (kernel size 3, stride 1), a group normalization layer (32 groups), a ReLu activation function and a MaxPool 2x2 operation. There are 64 channels for the first convolution and the number of channels is multiplied by 2 at each stack until reaching 512 for the 4th and last stack convolution. We then connect our 8x8x512 ultimate feature map with 2 fully-connected layers of size 1024 and 512 neurons and then our final output neuron on which we apply a binary cross entropy loss. We jointly trained our generator and discriminator on 50,000 images for 40 epochs. Training is performed on batches of size 8 with the Adam optimizer. Given our small training batch size, we found the use of group norm \cite{Wu2018GroupNorm} to be a great alternative to batch norm that was producing training instabilities. The learning rate is kept constant at 1e-4 during the first 20 epochs and is then decreased linearly to zero during the following 20 epochs. In practice, since our $\mathcal{L}_{L1}$ loss is much smaller than the $\mathcal{L}_{GAN}$ loss, we multiply the $\mathcal{L}_{L1}$ loss by a $\lambda$ factor equal to 1e4. With this adversarial training, we observed that the results are sharper and more realistic. In cases of deformed or missing limb, e.g. the legs in Figure~\ref{GAN_figure} right, the use of a discriminator forces the generator to produce a better prediction.

\section{Dataset generation}
\label{sec:dataset}

We introduce  3D HUMANS (HUman Motion, Activities aNd Shape), a realistic large-scale dataset  of humans in action with ground-truth 3D data (shape and pose). It consists of semi-synthetic videos with 3D pose and 3D body shape annotations, as well as 3D detailed surface including cloths and manipulated objects. First, we captured 3D meshes of humans in real-life situations using a multi-camera platform. We then rendered these models on real-world background scenes. See examples in Figure~\ref{dataset}a.

\noindent {\bf Capture.} We employed a state of the art 3D capture equipment
with 68 color cameras to produce highly detailed shape and appearance
information with 3D textured meshes. The meshes are reconstructed frame by frame
independently. They are not temporally aligned and do not share any common
topology. We divided the capture into 2 different subsets: in the first one, 13
subjects (6 male and 7 female) were captured with 4 different types of garments
(bathing suit/tight clothing, short/skirt/dress, wide cloths and jacket) while
performing basic movements e.g., walk, run, bend, squat, knees-up, spinning. In the second subset, 6 subjects, 4 male and 2
female, were captured while performing 4 different activities (talking on the
phone, taking pictures, cleaning a window, mopping the floor) in 2 different
ways: standing/sitting for talking on the phone, standing/kneeling for taking
picture, etc.  More than 150k meshes were reconstructed. The dataset was
collected at Inria from consenting and informed participants.

\noindent {\bf Rendering.} We rendered all our videos at a 320 x 240 resolution using a camera of sensor size 32mm and focal length 60mm. Our videos are 100 frames in length and start with the subject at the center of the frame. For the first frame of the sequence, the subject is positioned at a distance of 8 meters of the camera, with a standard deviation of 1 meter. We used the images of the LSUN dataset \cite{Yu15LSUN} for background. 

\noindent {\bf  Annotations.} We augment our dataset with ground-truth SMPL pose and body parameters. To do so, we use the Human3.6M~\cite{IonescuPOS14} environment as a ``virtual MoCap room'': we render the 3D meshes for which we want to estimate the 3D pose within that environment, generate 4 views using camera parameters and background images from the dataset and estimate the 2D/3D poses by running LCR-Net++, an off-the-shelf 3D pose detector particularly efficient on Human3.6M. An optimum 3D pose is then computed using multi-view 3D reconstruction and used as initialization to fit the SMPL model, estimating pose and shape parameters that better match each mesh. The SMPL model is fitted to the point clouds both for naked and dressed bodies.   Keeping the body parameters fixed (obtained from fits in minimal clothing) resulted in a lower performance of the baseline when evaluated against ground truth dressed bodies.

\begin{figure}[htp]
 	\centering
	\begin{tabular}{cc}
	\hspace{-5mm}
   	\includegraphics[width=.25\textwidth]{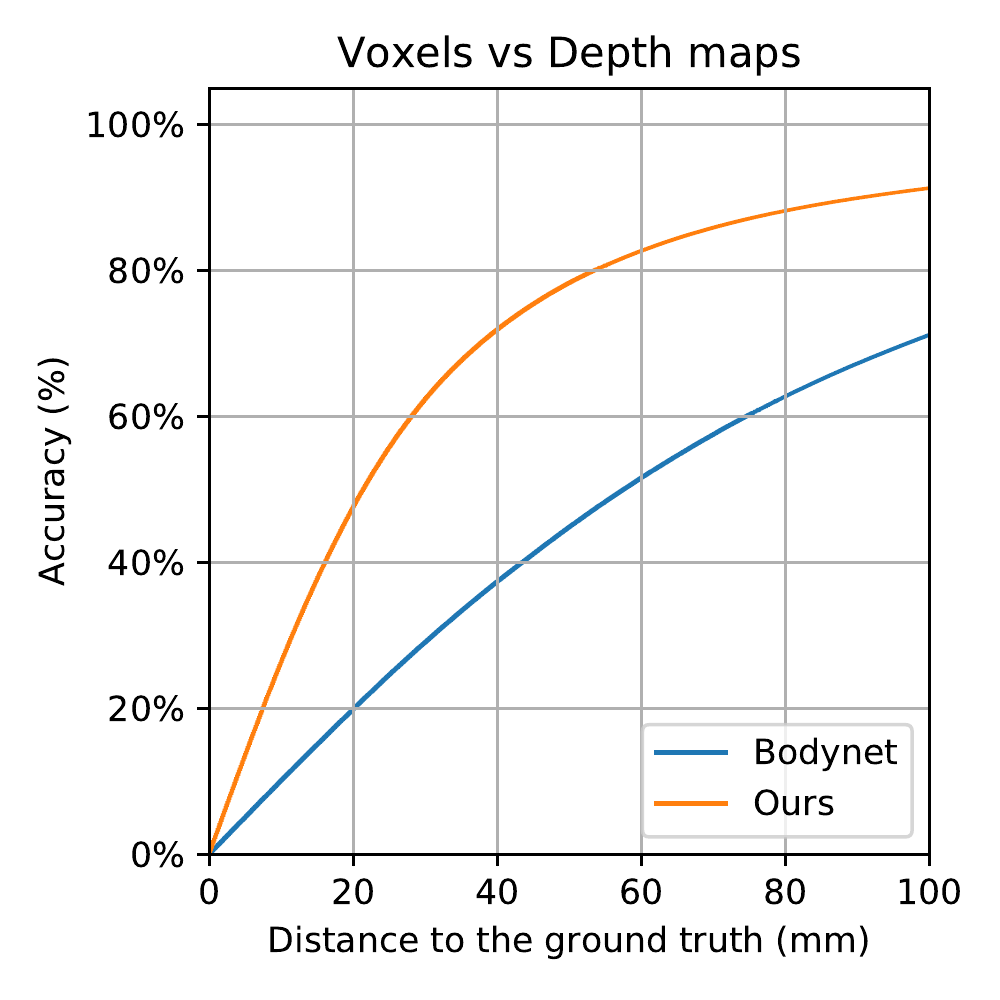}&
	\hspace{-5mm}
 	\includegraphics[width=.25\textwidth]{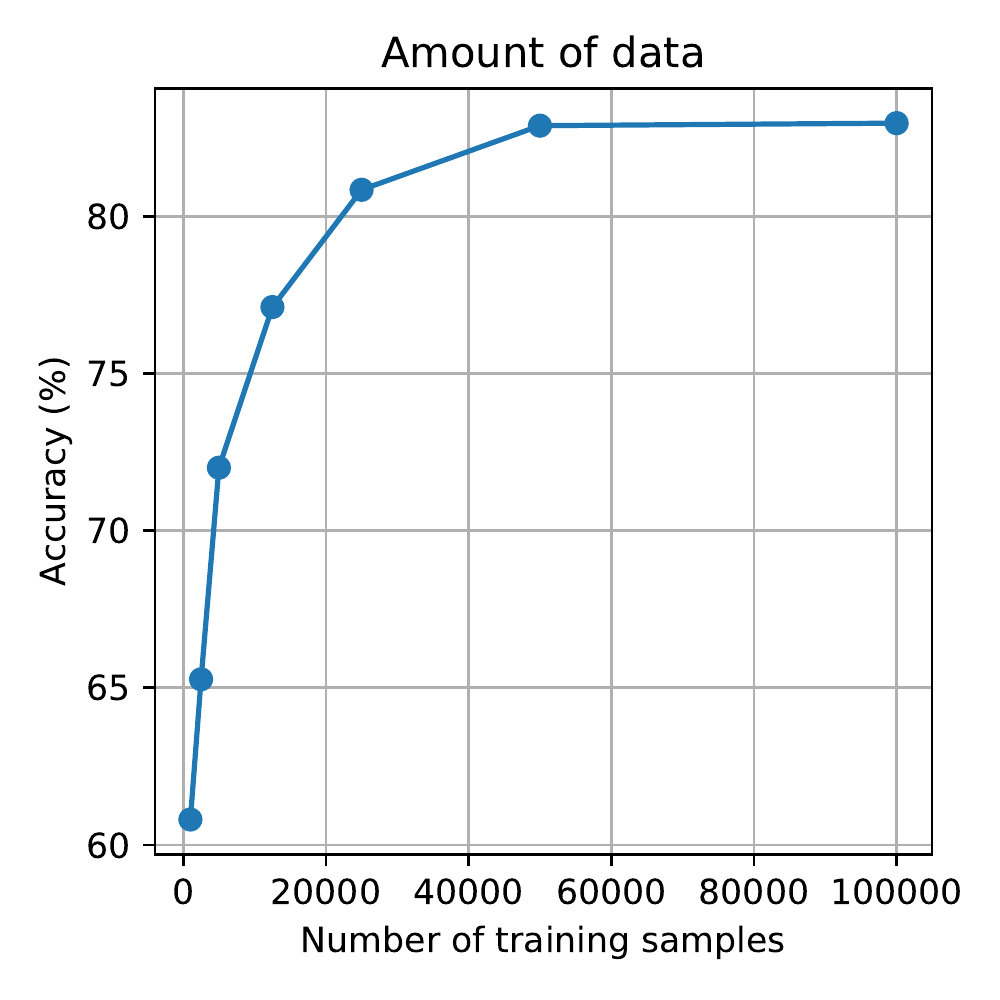}\\
	(a)&(b)
	\end{tabular}
   	\caption{Comparaison with state-of-the-art on the SURREAL dataset: (a) we first  compare against the BodyNet~\cite{Varol2018BodyNet} baseline. (b) We analyse the impact of varying the size of the training set on performance on our new 3D HUMANS dataset.}
   	\label{surreal}
	\vspace{-5mm}
\end{figure}

\section{Experiments}
\label{sec:xp}   
 
 We analyse quantitatively and compare our approach to the state-of-the-art on two datasets. First, the SURREAL dataset~\cite{Varol2017Surreal}, a synthetic dataset obtained by animating textured human models using MoCap data and rendering them on real background images, and our 3D HUMANS dataset introduced in this paper. While SURREAL covers a wider range of movements since it has been rendered using thousands of sequences from~\cite{CMUposedataset}, our data better covers shape details such as hair and clothing. %
In the following experiments, both training and test images are tightly cropped around the person using subjects segmentation. The smallest dimension of the image is extended to obtain a square image that is then resized to 256x256 pixels to serve as input for our network. 
Performance is computed on both 128x128 output depth maps as the distance
between each ground truth foreground pixel and its corresponding pixel in the
predicted depth map. Background depth $L$ is set at 1.5m.

\begin{figure*}[htp]
 	\centering
	\begin{tabular}{cccc}
	\hspace{-5mm}
 	\includegraphics[width=.25\textwidth]{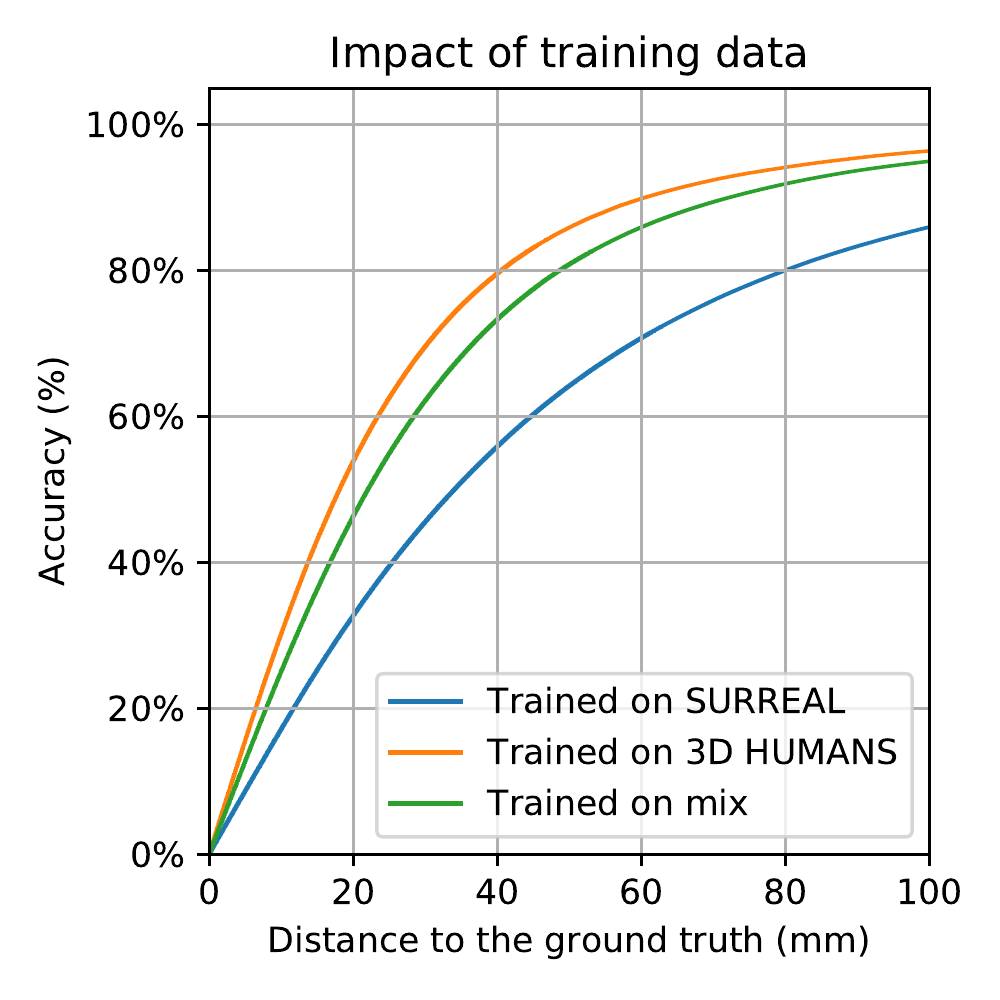}&
	\hspace{-5mm}
   	\includegraphics[width=.25\textwidth]{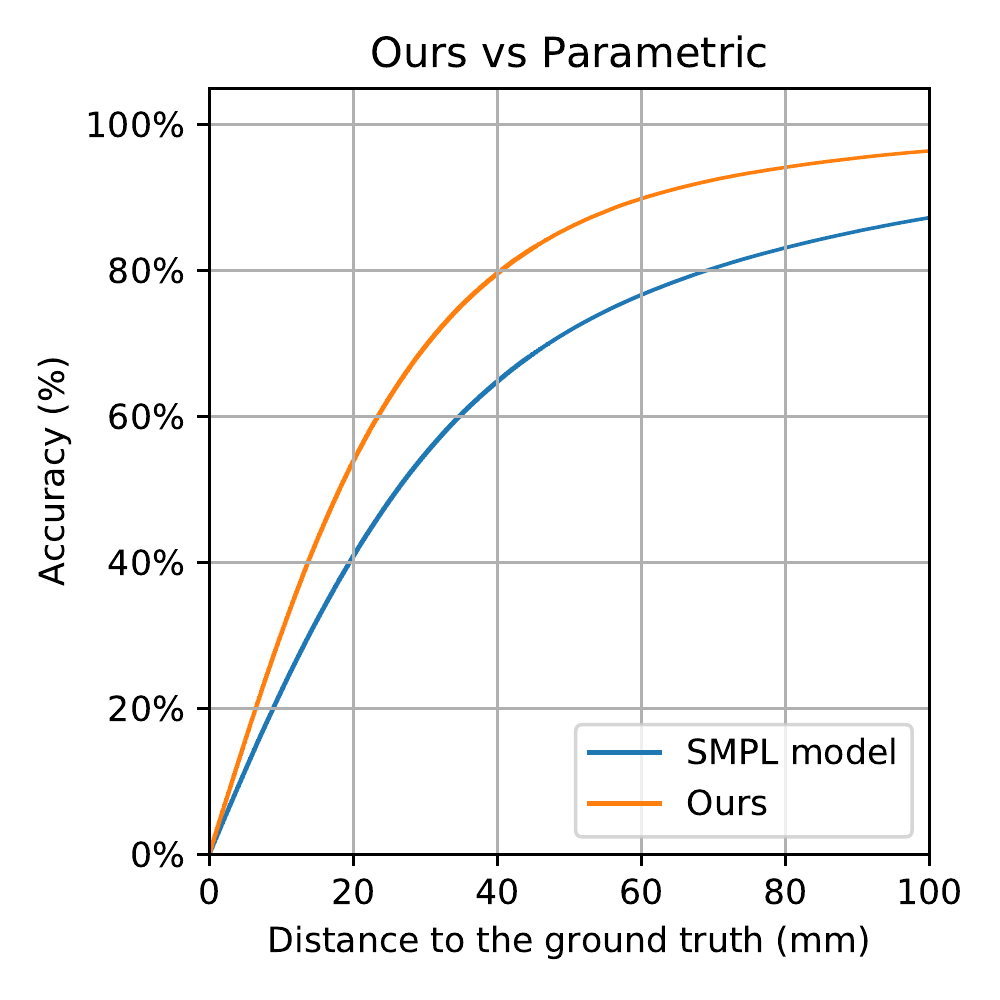}&
	\hspace{-5mm}
  	\includegraphics[width=.25\textwidth]{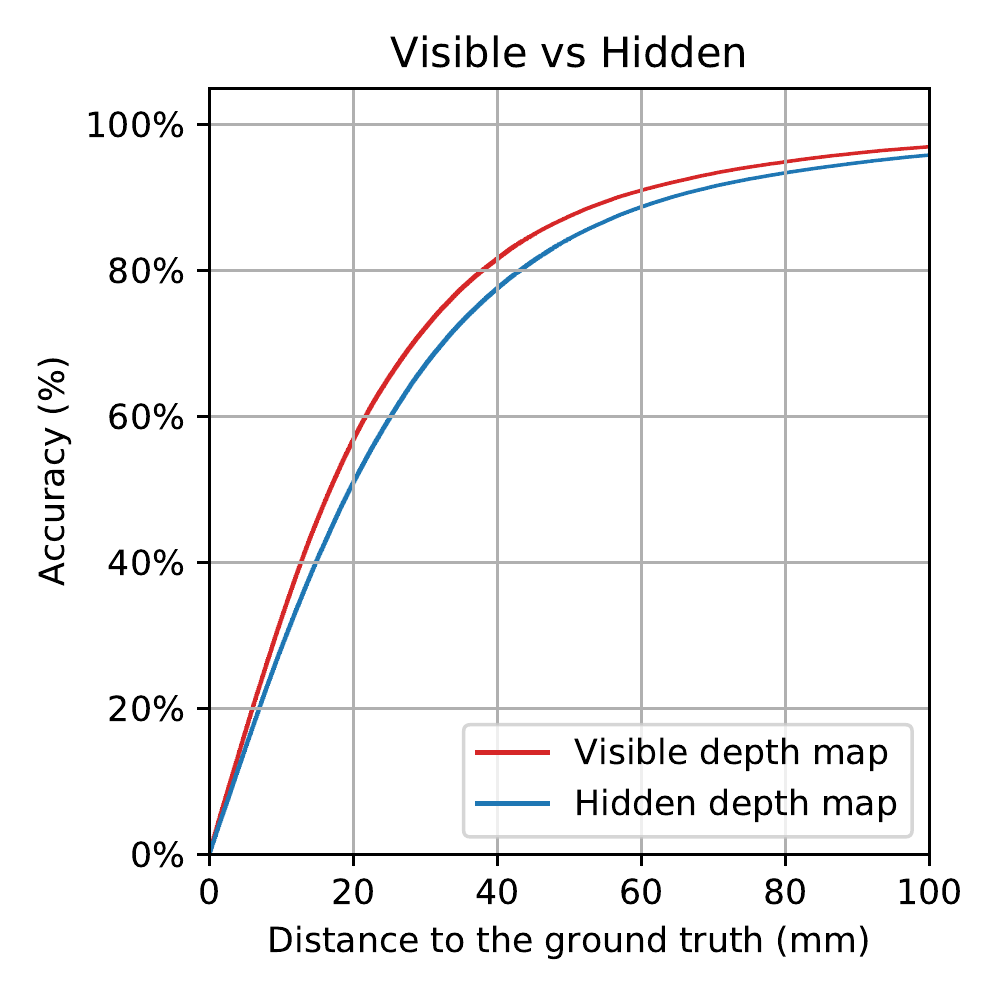}&
	\hspace{-5mm}
	\includegraphics[width=0.25\textwidth]{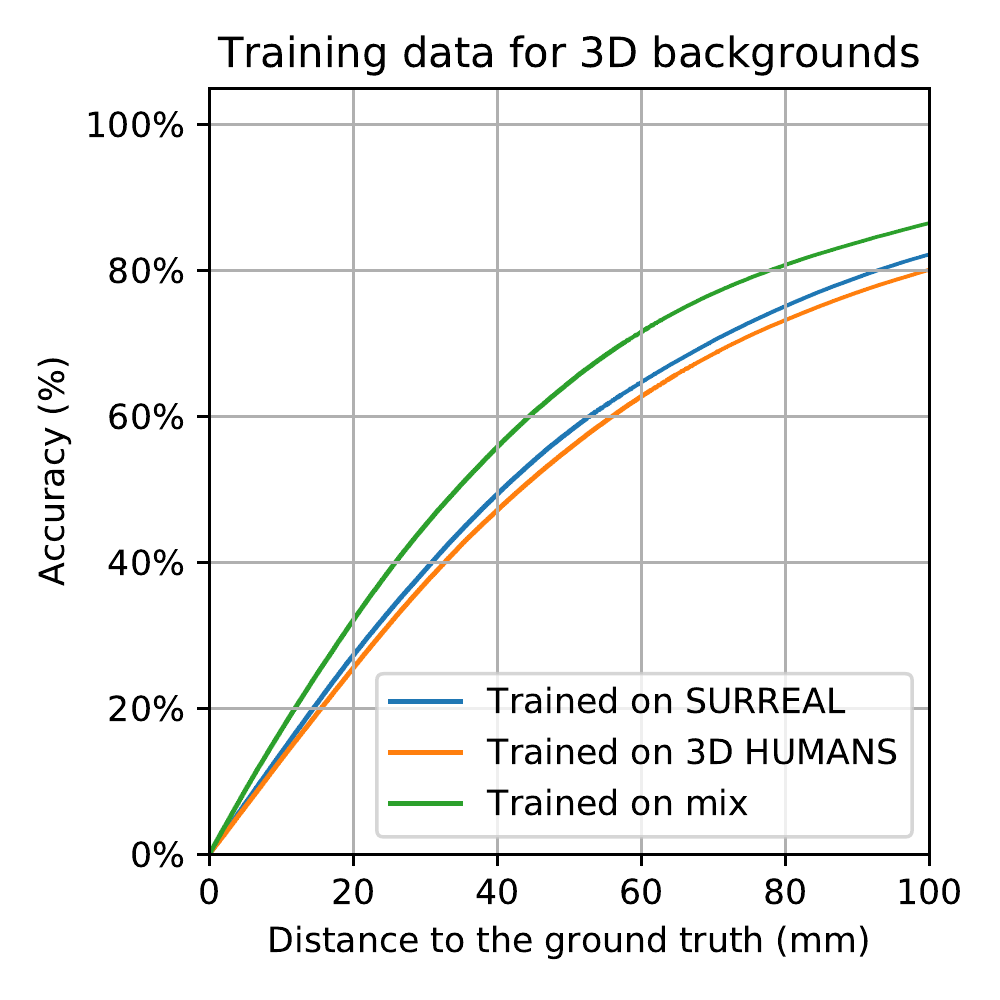}\\
	(a)&(b)&(c)&(d)
	\end{tabular}
   	\caption{Evaluation on our 3D HUMANS dataset: we first analyze the influence of the training data on performance (a). Then, we compare against the SMPL baseline (b). We compare the performance on visible and hidden depth map separately (c). Finally, we analyse the training data on a dataset rendered in realistic backgrounds and observe that SURREAL data is important for generalisation (d).}
   	\label{humans}
	\vspace{-3mm}
\end{figure*}

\subsection{SURREAL}

Recent methods~\cite{Varol2018BodyNet,Varol2017Surreal} evaluate their performance on this dataset. 
First, we evaluated the performance of our architecture when estimating quantized depth values (19+1 for background) through classification as in~\cite{Varol2017Surreal} and our proposed regression method: with a maximum distance to groundtruth of 30mm, the quantity of pixels with a correct depth estimation increase by $5\%$ when using regression instead of classification. Then, we compare in Figure~\ref{surreal}a our performance against the recent  BodyNet voxel grid-based architecture from~\cite{Varol2018BodyNet} who also reported numerical performance on SURREAL. Although good 3D performances are reported in the paper, we can see that when evaluating in the image domain, i.e., comparing depth maps, the performance of BodyNet drops. Our method makes 3D estimations aligned with the image and better recover details, outperforming BodyNet quite substantially.

\begin{figure}[h]
\centering
\includegraphics[width=\columnwidth]{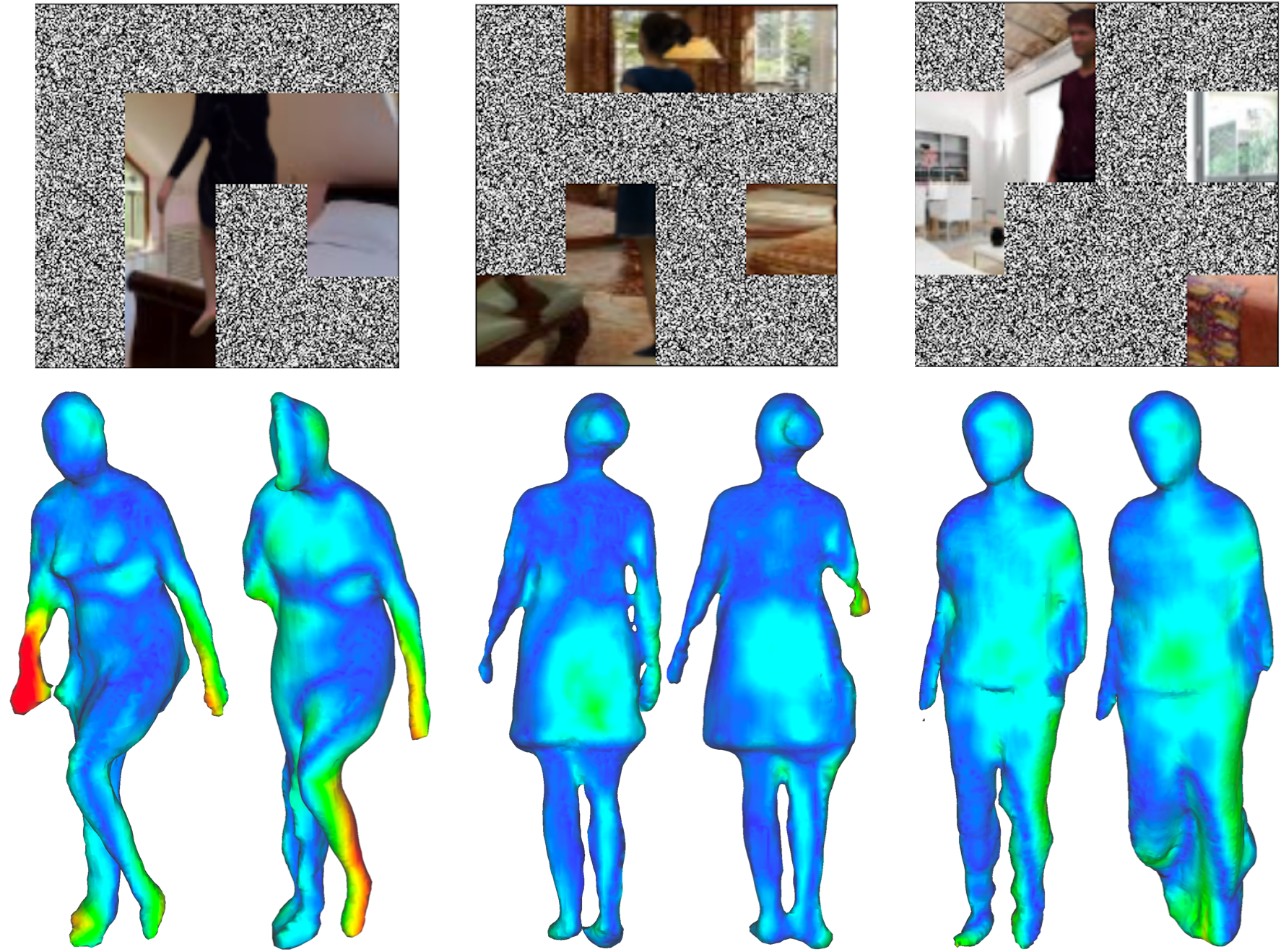}
\caption{Performance on 3D-HUMANS dataset in presence of severe occlusions on three frames: (top) input images, (left) with GAN, (right) without GAN. Errors above 15cm are shown in red. The GAN helps increase the ``humanness'' of the predictions.}
\label{GAN_figure} 
\end{figure}

\subsection{3D HUMANS}

We consider 14 subjects (8 male, 6 female) for training and the remaining 5 subjects (2 male, 3 female) for test.
An interesting aspect of synthetic datasets is that they offer an almost unlimited amount of training data. In our case, the data generation relies on a capture process with a non-negligible acquisition effort. It is therefore interesting to analyze how adding more training data impacts the performance. Our results in Figure~\ref{surreal}b show that training our architecture on 50,000 images is sufficient and that using more training images does not improve much the performance. 
The appearance of our images being quite different from SURREAL data, we first compare the performance of our method when considering different training strategies: training on SURREAL, training on 3D HUMANS, or training on a mix of both datasets. In Figure~\ref{humans}a, we can see that the best performances are obtained when SURREAL images are not used. The appearance of the images is too different and our architecture cannot recover details such as clothes or hair when trained on data obtained by rendering the SMPL model. This is verified by the result depicted in Figure~\ref{humans}b: we outperform, by a large margin, a baseline obtained by fitting the SMPL model on the ground
truth meshes, effectively acting as an upper bound for all methods
estimating SMPL meshes~\cite{Bogo2016SMPLify,KanazawaBJM18}.  It shows the
inefficiency of these methods to estimate clothed body shape since
clothes are not included in the SMPL model.

\begin{figure}[ht]
\centering
\includegraphics[width=\columnwidth]{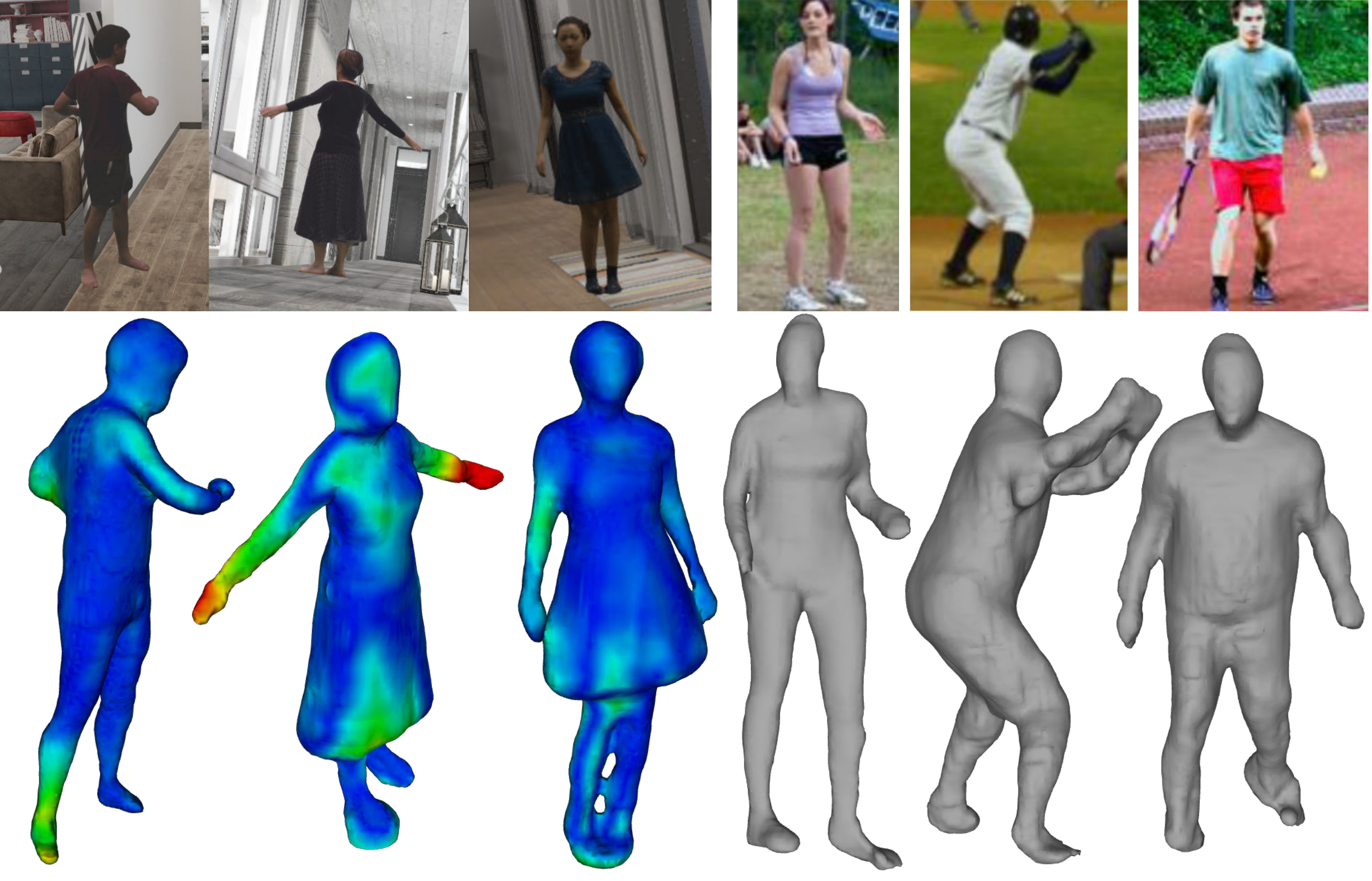}
\caption{Generalisation to previously unobserved data. We apply our pipeline to images with 3D realistically rendered backgrounds (left), and with 3 real-world images  from the LSP dataset (right). These poses, in particular the baseball player, have not been seen at training time but our model still generalizes well.}
\label{qualitatif} 
\vspace{-5mm}
\end{figure}

Finally, we analyse in Figure~\ref{humans}c how much our performance varies between front and back depth maps. As expected, we better estimate the visible depth map, but our hidden depth maps are usually acceptable. See  examples in Figure~\ref{GAN_figure} and Figure~\ref{qualitatif}. The quality of the 3D reconstructions is remarkable given the low dimensionality of the input. Main failures occur when a limb is completely occluded. In such cases, the network can create non-human shapes. We proposed to tackle this issue by considering an adversarial training that we analyse in the next section.
We note a higher performance on 3D HUMANS than on SURREAL. We attribute that to several factors including the higher pose variability in SURREAL (some subjects are in horizontal position) and the absence of lighting in 3D HUMANS. We also analyzed the results on different subsets of the evaluation set @50mm and obtained with/without clothing: 83.30$\%$ and 85.43$\%$ respectively and with/without object: 79.11$\%$ and 84.55$\%$ respectively, confirming the nuisance introduced by these elements.

\subsection{GAN}

Severe occlusions (self- or by other elements of the scene)  are a limitation of
our model that we address with adversarial training. We carried out a dedicated
experiment where we artificially generated such occlusions in train/test images
to quantify improvements. We obtain a 7\% chamfer distance error drop with adversarial training and a clear qualitative improvement which we illustrate in Figure~\ref{GAN_figure}. We highlight the differences by showing an error heat-map over a Poisson reconstruction of the point cloud for better visualization. The quantitative gain is limited due to the network sometimes hallucinating plausible limbs far from groundtruth (red hand in the left Figure~\ref{GAN_figure}), resulting in higher error than a network without GAN that does not estimate any limb at all. This is because the metric does not evaluate the overall plausibility of the produced estimation.

\subsection{Generalisation}

\begin{figure}[ht]
	\centering
	\includegraphics[clip, trim=0cm 0cm 0cm 1cm,width=1\columnwidth]{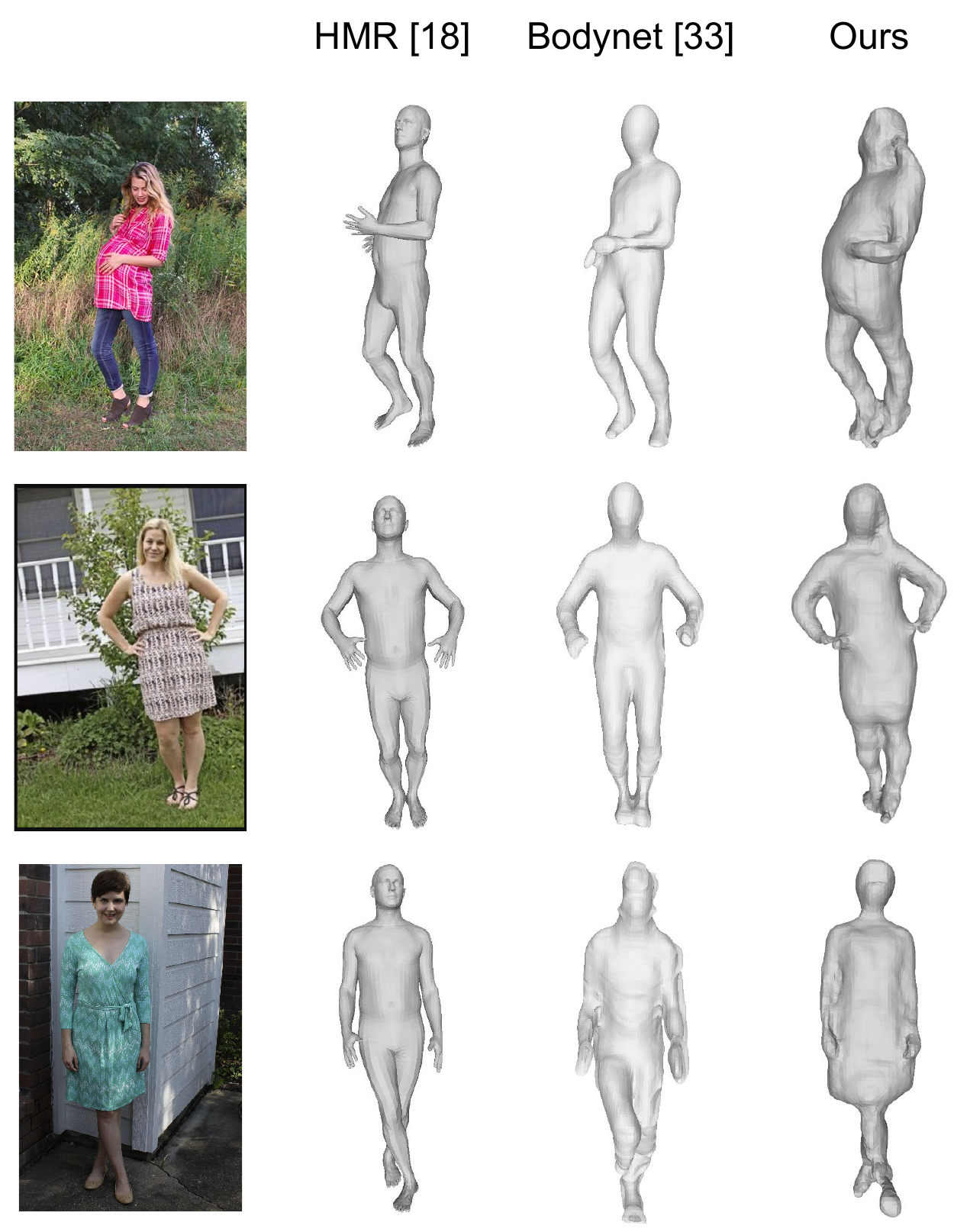} 
  \caption{Comparison between HMR~\cite{KanazawaBJM18} (left), Bodynet~\cite{Varol2018BodyNet} (middle) and our method (right). Unlike~\cite{KanazawaBJM18,Varol2018BodyNet}, we do not train on in-the-wild images but  estimate 3D shapes of clothed subjects. }
	\label{visual_comparison}
	\vspace{-6mm}
\end{figure}

In order to quantitatively measure its generalisation capability, we have evaluated our network on an additional dataset: instead of static background images, we have rendered the meshes in realistic 3D environments obtained on the internet (examples in Figure~\ref{dataset}b). The results (Figure~\ref{humans}d) show that a mix training on both SURREAL and 3D HUMANS is ideal for generalisation. We suspect that jointly rendering the subject and the 3D background at the same time creates a more realistic image where the subject is more complicated to segment, hence the need for more variability in the training data.
We also generated qualitative results for LSP images~\cite{Johnson10}, depicted in Figure~\ref{qualitatif}, and for the DeepFashion dataset~\cite{LiuLQWT16}, shown
 in Figure~\ref{visual_comparison} where we compare
our approach with HMR~\cite{KanazawaBJM18} and BodyNet~\cite{Varol2018BodyNet}.  We can observe that our approach captures more details, including hair, shirt and the belly of the pregnant woman (up), hair, skirt and body pose (middle) and dress (bottom).

\section{Conclusion}
\label{sec:conclusions}

We have proposed a new non-parametric approach to encode the 3D shape of a person through a double 2.5D depth map representation: a ``visible'' depth map depicts the elements of the surface that are directly observable in the image while a ``hidden'' depth map characterises the occluded 3D surface. We have designed an architecture that takes as input a single image and simultaneously produces an estimate for both depth maps resulting, once combined,  in a point cloud of the full 3D surface. Our method can recover detailed surfaces while keeping the output to a reasonable size. This makes the learning stage more efficient. Our architecture can also efficiently incorporate a discriminator in an adversarial fashion to improve the accuracy and ``humanness'' of the output.  To train and  evaluate  our network, we have captured a large-scale dataset of textured 3D meshes that we rendered on real background images. This dataset will be extended and released to spur further research.

\vspace{3mm}
\noindent {\bf Acknowledgements.}  We thank Pau De Jorge and Jinlong Yang for their help in capturing the data employed in this paper. The dataset was acquired using the Kinovis\footnote{https://kinovis.inria.fr} platform. This work was supported in part by ERC advanced grant Allegro.

{\small
\bibliographystyle{ieee}
\bibliography{egbib}
}

\vspace{10mm}
\noindent{\textbf{\LARGE Annex}}\\

\section{Architecture details}
\label{sec:architecture}

\subsection{Generator}
The main difference between our generator architecture and the stacked
hourglass by Newell et al.~\cite{NewellY16StackedHourglass} is the output dimension.
Newell et al.~estimate a 64x64 resolution heatmap for each body joint.
In our case, we estimate 2 depth maps and aim at a higher 128x128 resolution. Our hourglass
output dimension is 128x128x2. 
Because of this difference in output resolution, we apply the following
modifications to the stacked hourglass~\cite{NewellY16StackedHourglass} architecture: We do not use a maxpooling
operation after layer1, we increase the depth of the hourglasses from 4 to 5 skipped
connections, we project the hourglass result on 2 channels (one for each depth map).
Also, we use 2 stacked hourglasses and we replace batch normalization by group
normalization \cite{Wu2018GroupNorm} that performs better on
small training batches. See architecture details in Table~\ref{tab:generator_architecture}.

\begin{center}
   \begin{table}[ht]
     \fontsize{8}{12}\selectfont
    \begin{tabularx}{\linewidth}{| l | X | l |}
    \hline
      {\bfseries Layer} & {\bfseries Layer type} & {\bfseries Output shape} \\ \hline
      Input & Input & 256x256x3 \\ \hline
      Conv1 & Conv 7x7 stride=2, GroupNorm, Relu & 128x128x64 \\ \hline
      Layer1 & Residual module expanded & 128x128x128 \\ \hline
      Layer2 & Residual module expanded & 128x128x256 \\ \hline
      Layer3 & Residual module & 128x128x256 \\ \hline
      Hg1 & Hourglass, skipped connections = 5 & 128x128x2 \\ \hline
      Hg2 & Hourglass, skipped connections = 5 & 128x128x2 \\ \hline
      
    \end{tabularx}
    \caption{\label{tab:generator_architecture}Generator architecture.}
    \end{table}
\end{center}

\subsection{Discriminator}

For our discriminator, we employed a 4 stacks CNN. It takes as input a set of 2
depth maps at resolution 128x128 and outputs a scalar: close to 1.0 if it believes they are sampled
from the ground truth depth maps and close to 0 if it believes they have been generated by
the generator. See
Table~\ref{tab:discriminator_architecture} for details.

\begin{center}
  \begin{table}[ht]
     \fontsize{8}{12}\selectfont
    \begin{tabularx}{\linewidth}{| l | X | l |}
    \hline
      {\bfseries Layer} & {\bfseries Layer type} & {\bfseries Output shape} \\ \hline
      Input & Input & 128x128x2 \\ \hline
      Conv1 & Conv 3x3 stride=1, GroupNorm, Relu & 128x128x64 \\ \hline
      MP1 & MaxPool 2x2 & 64x64x64 \\ \hline
      Conv2 & Conv 3x3 stride=1, GroupNorm, Relu & 64x64x128 \\ \hline
      MP2 & MaxPool 2x2 & 32x32x128 \\ \hline
      Conv3 & Conv 3x3 stride=1, GroupNorm, Relu & 32x32x256 \\ \hline
      MP3 & MaxPool 2x2 & 16x16x256 \\ \hline
      Conv4 & Conv 3x3 stride=1, GroupNorm, Relu & 16x16x512 \\ \hline
      MP4 & MaxPool 2x2 & 8x8x512 \\ \hline
      FC1 & Fully connected layer & 1024 \\ \hline
      FC2 & Fully connected layer & 512 \\ \hline
      FC3 & Fully connected layer & 1 \\ \hline
      
    \end{tabularx}
    \caption{\label{tab:discriminator_architecture}Discriminator architecture.}
    \end{table}
\end{center}

\end{document}